\documentclass[sigconf]{cidr-2025}

\usepackage{multirow}
\usepackage{enumitem}
\usepackage{listings}
\usepackage{caption}
\usepackage{subcaption}

\newcommand{\tab}[1]{Table~\ref{#1}}
\newcommand{\fig}[1]{Fig.~\ref{#1}}
\newcommand{\sect}[1]{\S\ref{#1}}

\newcommand{\eg}{\textit{e.g.}, } 
\newcommand{\ie}{\textit{i.e.}, }

\newcommand{\genedit}{\textsc{GenEdit}}

\usepackage{color}

\begin{document}

\title[\genedit:~Compounding Operators and Continuous Improvement to Tackle Text-to-SQL in the Enterprise]{GenEdit: Compounding Operators and Continuous Improvement to Tackle Text-to-SQL in the Enterprise}

\author{Karime Maamari}
\affiliation{%
	\institution{Distyl AI}
	\city{}
	\country{}
}
\email{karime@distyl.ai}

\author{Connor Landy}
\affiliation{%
	\institution{Distyl AI}
	\city{}
	\country{}
}
\email{connor@distyl.ai}

\author{Amine Mhedhbi}
\affiliation{
	\institution{Polytechnique Montreal}
	\city{}
	\country{}
}
\email{amine.mhedhbi@polymtl.ca}

\begin{abstract}
Recent advancements in Text-to-SQL, driven by large language models, are democratizing data access. 
Despite these advancements, enterprise deployments remain challenging due to the need to capture business-specific knowledge, 
handle complex queries, and meet expectations of continuous improvements. 
To address these issues, we designed and implemented \genedit: our Text-to-SQL generation system that improves with user feedback. 
\genedit~builds and maintains a company-specific knowledge set, employs a pipeline of operators decomposing SQL generation, and uses feedback to update its knowledge set to improve future SQL generations. 


We describe \genedit's architecture made of two core modules: 
(i) decomposed SQL generation; and 
(ii) knowledge set edits based on user feedback. 
For generation, \genedit~leverages compounding operators to improve knowledge retrieval and to create a plan as chain-of-thought steps that guides generation. 
\genedit~first retrieves relevant examples in an initial retrieval stage where original SQL queries are decomposed into sub-statements, clauses or sub-queries. It then also retrieves instructions and schema elements. 
Using the retrieved contextual information, 
\genedit~then generates  step-by-step plan in natural language on how to produce the query. 
Finally, \genedit~uses the plan to generate SQL, minimizing the need for model reasoning, which enhances complex SQL generation. If necessary, \genedit~regenerates the query based on syntactic and semantic errors. 
The knowledge set edits are recommended through an interactive copilot, allowing users to iterate on their feedback and to regenerate SQL queries as needed. Each generation uses staged edits which update the generation prompt.  
Once the feedback is submitted, it gets merged after passing regression testing and obtaining an approval, improving future generations. 
\end{abstract}


\maketitle

\section{Introduction}

Recent advancements in Text-to-SQL have broadened access to data for a wider range of users while enabling faster query authoring using database management systems (DBMSs). 
At the core of these advancements are large language models (LLMs), 
which achieve unprecedented accuracy through in-context learning (ICL)~\cite{DBLP:conf/nips/PourrezaR23,DBLP:journals/corr/abs-2310-17342} and fine-tuning~\cite{DBLP:journals/pvldb/GaoWLSQDZ24,10.1145/3654930,DBLP:journals/corr/abs-2408-07702/death-schema}. 
Nevertheless many of these approaches are not readily deployable in the enterprise setting. 
From our experience with customer deployments, a Text-to-SQL solution ought to: 
i) understand \emph{external knowledge}, \eg a company's specific terminology and processes; 
ii) handle \emph{large query complexity} due to the schemas of data warehouses and the inherent complexity of the queries themselves; and 
iii) improve \emph{systematically over time}. 

Consider the following query, denoted as $Q_{fin-perf}$, which will serve as a running example. 
This query originates from a data analyst working at a holding company with ownership in multiple sports organizations.\footnote{While $Q_{fin-perf}$ represents the actual structure of a query in production, we mask the domain and any relevant customer or user details.}
The query is defined as follows:
\begin{lstlisting}[frame=single]
Identify our 5 sports organisations with the 
best and worst QoQFP in Canada for Q2 2023.
\end{lstlisting}
The query asks for QoQFP, \ie quarter-over-quarter financial performance. 
The QoQFP acronym within this company has a very specific meaning and associated calculations; 
the query cannot be answered without understanding such specifics. 
An LLM might in fact have a different interpretation of this acronym due to its pre-training. 
Furthermore, the query's equivalent 
SQL has very high complexity. 
The equivalent SQL to $Q_{fin-perf}$ is shown in Appendix~\sect{sec:output-sql-generated-by-language-model} with appropriate domain and data masking. 
Such highly complex queries have been reported as a core Text-to-SQL challenge~\cite{DBLP:journals/corr/abs-2406-08426/nextgen-db-interfaces-survey} and differ significantly to the common queries found in current popular public benchmarks~\cite{li2023llm,yu2019spider}. 
Finally, data analysts expect that even if the initial generated SQL fails, 
the query generation should improve over time. 

\genedit~is our purpose-built system to address the challenges of enterprise scenarios similar to the prior example. 
First, \genedit~captures the specific context of a company by building and maintaining a knowledge set. 
The set is a \emph{view} containing pairs of: i) natural language; and ii) SQL examples, natural language instructions (or hints) for generation, and database schemas. 
Second, \genedit~handles the complexity challenge by relying on a multi-operator pipeline using LLM calls to decompose the problem and generates a step-by-step plan thereby reducing the need for LLM reasoning. 
Finally, \genedit~improves over time as users provide free text feedback leading to editing suggestions to the knowledge set and prompts. 
This in turn improves future generations. 
\genedit~contains an edits recommendation module that supports subject matter experts in improving the system 
without understanding its internals. 

Previous work on Text-to-SQL shares similarities with our approach. 
For example, \genedit~pipelines are decomposed into operators and rely on few-shot examples for generation~\cite{DBLP:conf/nips/PourrezaR23,DBLP:conf/icde/RenFHHDHJZYW24,wang2024macsql}. 
Our approach however deviates in important ways. 
\genedit~operators do not retrieve or produce separate context fed to the LLM for generation; 
instead, they compound, \eg the choice of relevant examples informs the choice of instructions to retrieve. 
\genedit~also imposes a very different intermediate representation fed to the model to generate SQL. 
Instead of using full Text-to-SQL pairs as examples, \genedit~decomposes examples into smaller clauses, which are used for the chain-of-thought (CoT) optimization~\cite{DBLP:conf/nips/Wei0SBIXCLZ22,DBLP:journals/corr/abs-2310-17342}. 
\genedit~first generates a CoT plan in which one or more steps describe a Common Table Expression (CTE) expected to be part of the output candidate query. 
\genedit~then constructs its final output by combining these CTEs as described in the plan with an added SELECT-FROM-WHERE instruction.

In the rest of the paper, we overview \genedit's architecture~(\sect{sec:genedit-architecture}). 
We primarily cover our insights designing \genedit, which are:\vspace*{-0.5em}
\begin{itemize}[wide, 
	labelindent=0pt]
	\setlength{\itemsep}{3pt}
	\setlength{\parskip}{0pt}
	\setlength{\parsep}{0pt}
	\item \textbf{Compounding Operators} (\sect{subsec:compounding-operators}): 
	\begin{itemize}
		\setlength{\itemsep}{1pt}
		\setlength{\parskip}{0pt}
		\setlength{\parsep}{0pt}
		\item[$1$.] \emph{Context Expansion} (\sect{subsubsec:cascading}): we decompose SQL generation into multiple operators where the output of the prior operator is used in the subsequent one. 
		For example, consider a list of operators retrieving relevant contextual information for the input query from the knowledge set. 
		Let an initial operator select relevant examples of the task, \ie (natural language query, SQL) pairs. 
		The selection of these examples informs that of relevant instructions, which are natural language descriptions in the form of hints of what to do or definition of certain terms; the selection of both then informs the choice of relevant schema elements and so on. 
		Such context expansion is akin to \emph{query expansion} and improves the performance of subsequent retrieval operators by adding more relevant context to the query and hence finding more relevant elements in the knowledge set. 
		\item[$2$.] \emph{Planning for Generation} (\sect{subsubsec:planning-steps-for-generation}): in order to generate queries of high complexity, we minimize the need for LLM reasoning. 
		We do so by splitting the task of SQL generation into two 
		operators. 
		First, we generate an initial \emph{CoT plan}, 
		\ie a step-by-step natural language description of how to generate the output SQL query. 
		Each step describes a SQL sub-statement, a clause, or subquery in natural language and SQL. 
		Then, we use this plan and the retrieved knowledge as the context necessary to make a prediction. 
	\end{itemize}
	\item \textbf{Decomposing Examples} (\sect{subsec:decomposition-of-examples}): due to our novel CoT plan, instead of using few-shot examples as pairs of natural language and SQL queries, we decompose SQL examples repeatedly and have them at different granularity such as subqueries, clauses, or sub-statements. 
	The user's needs are best captured through the sub-statements obtained from such decomposition. 
	Many of the sub-statements end up being repeated across the space of expected SQL queries.
	As such, each step of the CoT plan describes generating a sub-statement similar to the examples. 
	\item \textbf{Recommending Edits} (\sect{subsec:editing-knowledge-set}): we found a large set of SQL generation issues to stem from either: 
	i) not understanding a particular context in the natural query (\eg use of `our' in the running example); 
	ii) a mistake in the decomposed examples (\eg how to do a specific calculation); or 
	iii) retrieval accuracy in the pipeline such as missing relevant schema elements or examples. 
	Hence, the primary approach to improve the system is to edit the knowledge set.
	An edit can update decomposed examples or instructions, insert new ones, or delete existing ones. 
	It can alternatively add instructions to the retrieval and reranking operations within the pipeline. 
	\item \textbf{Interactive Collaboration} (\sect{subsec:interactive-collaboration}): Text-to-SQL is not a stand-alone product and instead ships in our case within an analytics engine. 
	After a SQL generation, an SME can submit feedback and iterate on it through a user interface. 
	Submitting feedback generates recommendations of edits to the knowledge set. 
	The user chooses a subset if the edits to stage and regenerates the query. 
	Users can keep iterating on their feedback until they are satisfied with the regenerated query based on their staged edits. 
    The user then publishes the staged edits and can even make further direct edits themselves. 
	Once staged, the edits to the knowledge set are tested for regression. 
	Currently, these staged edits require human approval after passing regression testing. 
	All edits due to user feedback are logged into a history that can be audited and can be used to revert back to any prior checkpoint. 
\end{itemize}

\section{\genedit~Architecture}
\label{sec:genedit-architecture}


\begin{figure*}[t]
	\centering
	\captionsetup{justification=centering}
	\includegraphics[scale=0.8]{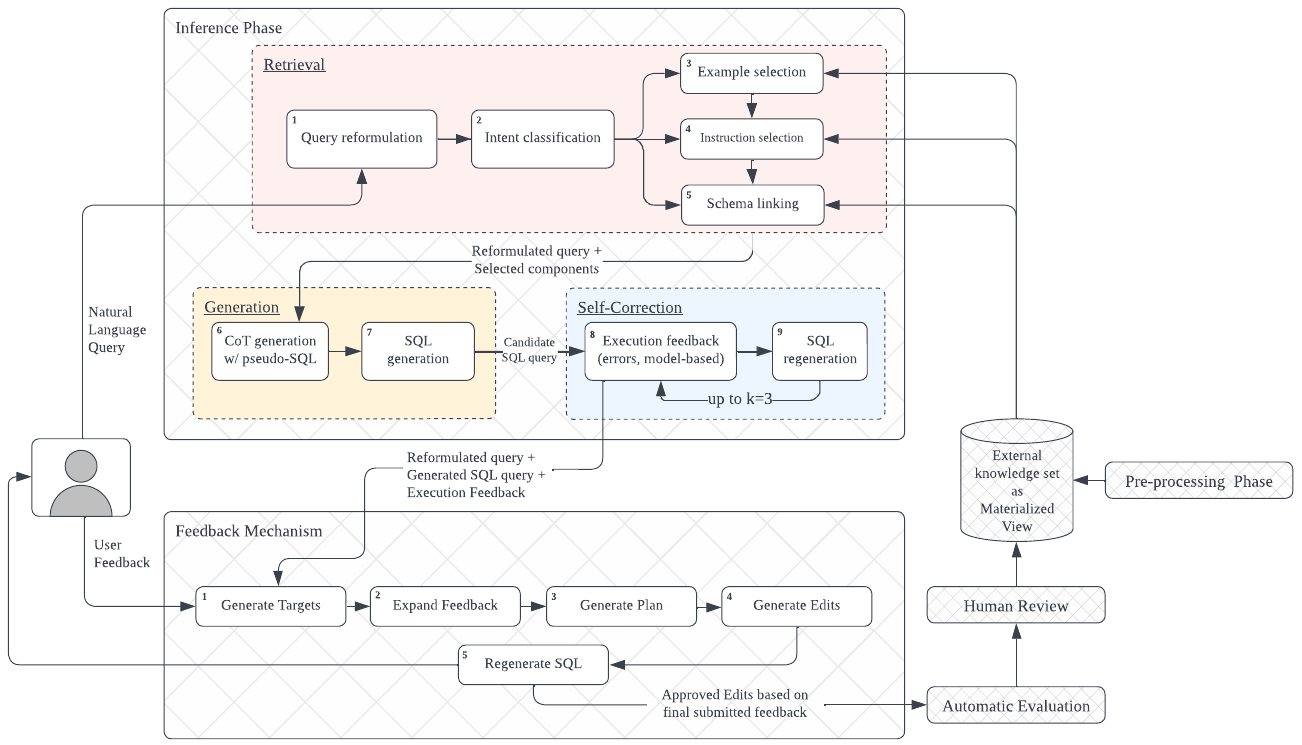}
	\caption{Overview of \genedit~architecture showcasing its pipeline operators for\\SQL generation and edits recommendation modules.}
	\Description[short description]{long description}
	\label{fig:architecture}
\end{figure*}

\genedit~has two core modules: a SQL generation module and an edits recommendation module. 
These are shown in the system architectural diagram in~\fig{fig:architecture}. 
The SQL generation components are highlighted by the \emph{pre-processing} and \emph{inference} phase boxes, 
while the edits recommendation module contains the \emph{feedback mechanism}, 
\emph{automatic evaluation}, and \emph{human review}. 
We give an overview of the SQL generation module and describe our ICL technique that has been deployed in production. 
The details of SQL generation and edit recommendation are in sections \sect{label:sec-generation} and \sect{sec:continuous-improvement}, respectively.

\subsection{SQL Generation Overview}
\label{subsec:sql-generation-overview}

We rely on pre-processing to construct the knowledge set. 
At inference, we use a subset of the relevant knowledge for generation: 
\begin{itemize}[wide, labelindent=0pt] 
	\setlength{\itemsep}{3pt}
	\setlength{\parskip}{0pt}
	\setlength{\parsep}{0pt}
	\item \textbf{\emph{Pre-processing}}\textbf{:} this phase takes as input: i) SQL queries from logs of prior executions; and ii) documents that contain domain-specific terminology and practices. 
	As output, it produces an external knowledge set as a materialized view that contains examples, instructions, and schema elements, grouped by \emph{user intents}. 
	A user intent describes a particular need or request, \eg \emph{`financial performance'} or \emph{`TV viewership numbers'} for $Q_{fin-perf}$. These intents are mined and verified by SMEs. 
	An important aspect of maintenance is keeping track of provenance in the view to update it as documents change.
	\genedit's specific representation for examples and instructions is novel (detailed in \sect{subsec:decomposition-of-examples}) and emerges in part from the decomposition approach at inference. 
	We will explain examples and instructions momentarily. 
	The schema is obtained directly from the database or documentation from data catalogs. 
	The schema is augmented with possible attribute values. Specifically, we add the top-5 most frequent values per attribute to the schema information. 
	\item \textbf{\emph{Inference}}\textbf{:} given a natural language input query, a pipeline made of multiple operators generates an output SQL query as follows. 
	First, \genedit~reformulates the query using a \emph{canonical format} (operator $1$ of inference in \fig{fig:architecture}) and classifies the query intents (operator $2$ of inference in \fig{fig:architecture}). 
	One example of changes to the query to conform to the canonical format is to always begin with ``Show me~...''. 
	Second, it retrieves the relevant knowledge from the view using the classified intents. It then relies on further retrieval operations of each component 
	to retrieve the next as shown in operators 3, 4, and 5 in the architectural diagram. 
	Finally, \genedit~generates a candidate query. 
	Unlike prior approaches, we decompose generation to minimize the need for LLM reasoning by performing two model calls. 
	The first call generates a step-by-step plan in natural language describing how to write the query and a second call uses the plan for the actual SQL generation. 
	In case of an error, \genedit~retries generation by adding the perceived errors as context and using self-correction similar to prior work~\cite{wang2018robust}. 
\end{itemize}

\section{SQL Generation}
\label{label:sec-generation}

In order to generate a candidate SQL query, \genedit~follows the inference phase in \fig{fig:architecture}. 
Given an input query, the compounding retrieval operators and the generation planning step help 
in constructing a prompt similar in structure to the one shown in \fig{fig:example-prompt}. 
This prompt is then used to produce one or more candidate SQL queries. 
If more than one candidate query is generated, \genedit~picks the 
\emph{`best'} one. \genedit~then might regenerate the query up to $k$ times based on syntactic and semantic errors. 

We cover the details of the retrieval operators with context expansion and the CoT planning prior to the SQL generation. We then cover the representation used for the examples and instructions. We note that the schema representation is similar to prior approaches and therefore is omitted~\cite{DBLP:journals/corr/abs-2406-12104}. 

\subsection{Compounding Operators}
\label{subsec:compounding-operators}

At inference, the pipeline classifies the intents of the input query and retrieves relevant context for generation. 
Each context retrieval operator relies on the output of prior ones specifically for re-ranking. 
We refer to this as context expansion which enables better re-ranking for the retrieval operators. 
In this section, we also cover how \genedit~decomposes SQL generation through planning, which in turn minimizes the need for LLM reasoning and allows us to tackle queries of much higher complexity. 

\subsubsection{Context Expansion}
\label{subsubsec:cascading}

Before retrieval, 
the query reformulation operator 
(\#1 in \fig{fig:architecture}) reformulates the query into a chosen canonical form. Then, the intent classification operator (\#2 in \fig{fig:architecture}) identifies the user intents. Recall that user intents are mined in a pre-processing phase and that each intent is associated with examples, instructions, and a set of schema elements, \ie tables and columns considered relevant to the intent. 


As such, the example selection operator (\#3 in \fig{fig:architecture}) uses the user intents to retrieve their associated examples from the view. It then retrieves further relevant examples based on the query.
The operator then re-ranks all examples based on a cosine similarity score with the reformulated query. 
Next, the instruction selection operator (\#4 in \fig{fig:architecture}) retrieves instructions in a similar fashion to the example selection operator; however, it re-ranks using not just the query but also the selected examples. 
Finally, the schema linking operator (\#5 in \fig{fig:architecture}) uses LLM calls to identify relevant elements as done in prior work~\cite{talaei2024chess} and adds a re-ranker for filtering to manage the LLM context of the model generating SQL~\cite{DBLP:journals/corr/abs-2408-07702/death-schema}. 


\begin{figure*}[t!]
	\centering
	\captionsetup{justification=centering}
	\includegraphics[scale=0.39]{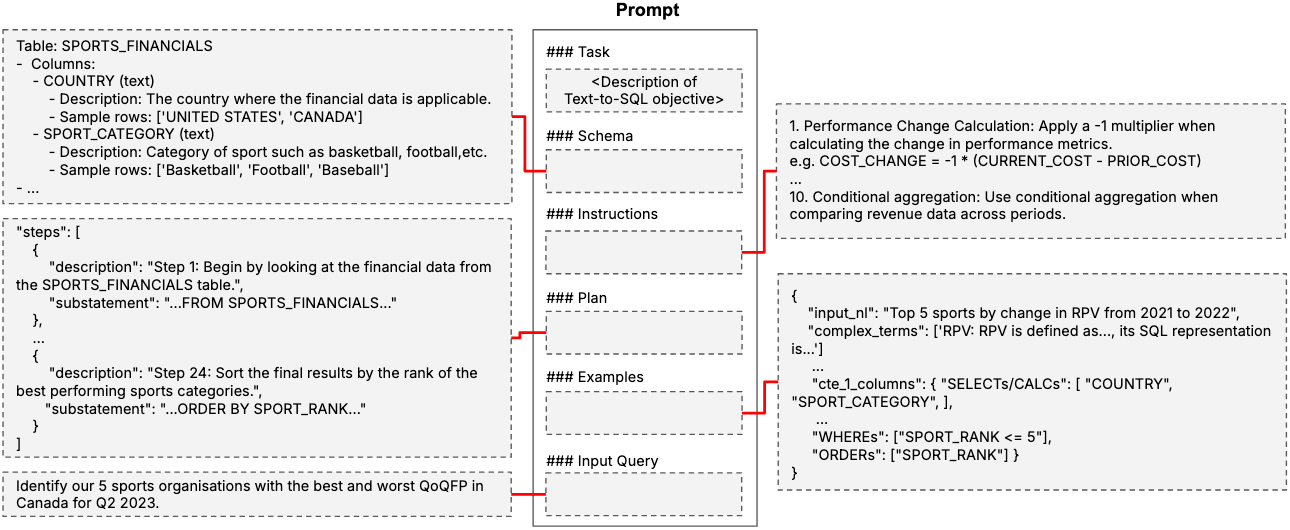}
	\caption{Example of retrieved knowledge and plan generated for  $Q_{fin-perf}$, which is then used for SQL prediction.}
	\Description[short description]{long description}
	\label{fig:example-prompt}
\end{figure*}

\vspace*{-0.3em}
\subsubsection{Planning for Generation}
\label{subsubsec:planning-steps-for-generation}

First, we construct a CoT reasoning plan using an LLM call with the reformulated input query and retrieved external knowledge~\cite{wei2023chainofthought, kojima2023large, wang2023planandsolve}.
The plan contains natural language steps of how to write the query. 
Each step in the plan is augmented with `\emph{pseudo-SQL}' inspired by selected examples. 
Pseudo-SQL refers to a SQL sub-statement associated with a step. The substatement is written surrounded by dots ($\dots$) as a prefix and a suffix indicating that it is part of a larger query. 
In the prompt, the CoT plan is represented as a JSON object containing an ordered list of steps where each element is a pair of step description in natural language and pseudo-SQL.
Consider the plan for $Q_{fin-perf}$ in \fig{fig:example-prompt}, its first step is \emph{``Begin by looking at the financial data from the SPORTS\_FINANCIALS table.''} with the pseudo-SQL: \texttt{``$\dots$FROM SPORTS\_FINANCIALS$\dots$''}. The plan in the example contains 24 steps in total.

We rely on another LLM call to generate one or more candidate SQL queries using the reformulated query, retrieved external knowledge, and CoT plan.


\subsection{Decomposing Examples}
\label{subsec:decomposition-of-examples}

In pre-processing, \genedit~creates a materialized view of examples and instructions using query logs and domain-specific documents. 
Below, we describe the actual representation of each component. 
\fig{fig:example-prompt} shows the structure of the various representation within a prompt for the running example query $Q_{fin-perf}$. 

\subsubsection{Example Representation}
Our examples are SQL sub-state\-ments and each has an associated equivalent natural language description. 
This is in contrast with traditional examples which are full SQL queries represented in their raw format. 

The sub-statements are obtained as follows. 
We take as input the full SQL queries from historical logs or directly from domain experts. 
We first rewrite the queries to use CTEs (WITH clause with subqueries). 
Then, each rewritten query is decomposed into sub-queries based on its subqueries in the WITH clauses, and finally into sub-statements based on inner clauses. 
As a result, we represent examples in the decomposed form as shown in \fig{fig:example-prompt}. The prompt shows various examples such as a CTE columns projection, a WHERE clause and an ORDER BY clause. 
We augment these sub-statements with natural language descriptions of the complex operations, tables, and sub-queries involved. 
For instance, in the example in \fig{fig:example-prompt}, we have the RPV (revenue per viewer) term defined and its SQL statement, where the RPV example is associated with an intent.  
Our use of CTEs in the output and associated decomposition is a novel approach to task decomposition, though it has inspirations from the prior sketch-based slot-filling techniques~\cite{xu2017sqlnet}. 

\subsubsection{Instruction Representation}
Instructions for SQL generations are natural language guidelines provided to the model detailing how to interpret the input and formulate it into a correct SQL query. 
We obtain the instructions from example queries and documents containing domain-specific terminology and practices. 
We represent them in natural language with expected SQL sub-expressions when relevant. 
For instance, \fig{fig:example-prompt} shows two instructions. The first explaining how to calculate performance change \emph{(`Apply a -1 multiplier when calculating the change in performance metrics')} and the second explaining the need to use conditional aggregations  \emph{(`when comparing revenue data across periods')}. 

\subsection{Evaluation}
\label{subsec:sql-generation-evaluation}

\subsubsection{Dataset.} We evaluate our framework on the BIRD benchmark~\cite{li2023llm}. 
BIRD is widely accepted as the most challenging Text-to-SQL benchmark available. It contains queries from 95 databases spanning multiple domains. 
The difficulty of BIRD is in the imprecision of its data, queries, and external knowledge, making the benchmark the most representative of the real-world Text-to-SQL problem to date. 
To simplify and minimize the evaluation cost, we use the dev set by sampling 10\% of each database, as proposed in the evaluation of prior work~\cite{talaei2024chess}.

\subsubsection{Evaluation Metric.} The BIRD benchmark relies on the 
the Execution Accuracy (EX) as an accuracy evaluation metrics. 
EX is the proportion of queries for which the output of the predicted SQL
query is identical to that of the ground truth SQL query. 

\begin{table}[t!]
	\centering
	\captionsetup{justification=centering}
	\begin{tabular}{rcccc}
		\hline
		\multirow{2}{*}{\textbf{Methods}} & \multicolumn{4}{c}{\textbf{BIRD-Dev}} \\ \cline{2-5}
		& \textbf{Simple} & \textbf{Moderate} & \textbf{Challenging} & \textbf{All} \\ \hline
		CHESS \cite{talaei2024chess} & 65.43 & 64.81 & 58.33 & 64.62 \\ \hline
		MAC-SQL \cite{wang2024macsql} & 65.73 & 52.69 & 40.28 & 59.39 \\ \hline
		TA-SQL \cite{qu2024generationalignitnovel}& 63.14 & 48.60 & 36.11 & 56.19 \\ \hline
		DAIL-SQL \cite{gao2023texttosql} & 62.5 & 43.2 & 37.5 & 54.3 \\ \hline
		C3-SQL \cite{dong2023c3} & 58.9 & 38.5 & 31.9 & 50.2 \\ \hline
		\hline
		\genedit & 69.89 & 39.29 & 36.36 & 60.61 \\ \hline
	\end{tabular}
	\caption{Performance of \genedit~on the BIRD benchmark, compared to prior solutions.}
	\label{table:1}
\end{table}

\subsubsection{Model Selection.} 

We use \texttt{GPT-4o} across all operators, except for schema linking, where we instead employ \texttt{GPT-4o-mini} to reduce primarily cost and then latency.

\subsubsection{Results.} 

\tab{table:1} showcases the performance of \genedit~aga\-inst prior approaches. 
Comparing accuracy on 10\% of the BIRD benchmark dev set, \genedit~ranks second compared to open-source submissions, \ie submissions with available code and tenth compared to all submissions at 60.91\%.\footnote{evaluation numbers from Aug. 2024} In \tab{table:2}, we present an ablation of various operators within our pipeline to showcase the benefit of each. We see instructions are providing the most benefit while examples the least. This does not minimize the usefulness of examples as they are what we use to add pseudo-SQL to the CoT plan, which provides the second most benefit. 

When comparing \genedit~with another approach we developed~\cite{DBLP:journals/corr/abs-2408-07702/death-schema}, we find that our newer approach outperforms \genedit~on the BIRD dev set with an EX of 67.21\%. 
The other approach has simpler operators then the one we describe here. 
It uses fine-tuned \texttt{GPT-4o} and maximizes the schema contextual information at generation. 
Most surprisingly however, we use \genedit~within our enterprise deployments as the other approach can't handle the same query complexity even though it outperforms \genedit~on BIRD.
Note that newer benchmarks such as Beaver~\cite{DBLP:journals/corr/abs-2409-02038/beaver} and Spider 2.0~\cite{lei2024spider20evaluatinglanguage} introduce more complexity to overcome such a mismatch between real deployments and public benchmarks. 

\begin{table}[t!]
	\centering
	\captionsetup{justification=centering}
	\begin{tabular}{lcccc}
		\toprule
		\textbf{Method} & \textbf{Sim.} & \textbf{Mod.} & \textbf{Chall.} & \textbf{Total} \\ \midrule\midrule
		\genedit & 69.89 & 39.29 & 36.36 & 60.61 \\ \hline
		w/o Schema Linking & 67.74 & 42.86 & 18.18 & 58.33 (↓ 2.28) \\ \hline
		w/o Instructions & 58.06 & 28.57 & 36.36 & 50.00 (↓ 10.61) \\ \hline
		w/o Examples & 69.89 & 35.71 & 9.09 & 59.09 (↓ 1.52) \\ \hline
		w/o Pseudo-SQL & 62.37 & 25.00 & 18.18 & 50.76 (↓ 9.85) \\ \hline
		w/o Decomposition & 66.67 & 46.43 & 18.18 & 58.33 (↓ 2.28) \\ \bottomrule
	\end{tabular}
	\caption{An ablation study looking at the EX change without (w/o) certain operators.}
	\label{table:2}
\end{table}

\section{Continuous Improvement}
\label{sec:continuous-improvement}

In our early deployments, SMEs would give feedback directly to engineers who would then update the knowledge set manually. 
This is an unsustainable approach that cannot scale across deployments nor as use cases are added. 
In this section, we overview our 
edit recommendations module, which primarily aims to shorten the feedback integration loop. 
\genedit~acts as a crowdsourcing system~\cite{DBLP:journals/cacm/DoanRH11} and as an interactive machine learning system~\cite{DBLP:journals/aim/AmershiCKK14}. 

Recommended edits to the knowledge set are generated using the provided feedback in a fully-automated manner, following feedback operators \#1 to \#4 in~\fig{fig:architecture}. 
These generated edits are reviewed by SMEs and staged within a UI. 
SMEs can regenerate and iterate on their feedback until satisfied. They can then submit their edits which are tested for regressions and upon approval published and merged. 
We explain the operators associated with generating recommended edits to the knowledge set and overview the module's UI. 

\subsection{Recommending Edits}
\label{subsec:editing-knowledge-set}

As shown in \fig{fig:architecture}, \genedit~takes a generated query and associated user feedback as input and generates edit recommendations that then lead to a query regeneration as output.

Recommending edits relies on 4 operators:\vspace*{-0.3em}
\begin{itemize}[wide, labelindent=0pt] 
	\setlength{\itemsep}{0pt}
	\setlength{\parskip}{0pt}
	\setlength{\parsep}{0pt}
	\item[i.] \textbf{Generate Targets}: operator \#1 determines which of the instructions and examples retrieved for generation are relevant to the user feedback. It also generates a brief explanation of why the feedback is indeed relevant.
	\item[ii.] \textbf{Expand Feedback}: operator \#2 expands on the prior explanation for why the feedback is relevant to the chosen elements (instructions and examples). 
	\item[iii.] \textbf{Planning of Edits}: operator \#3 takes the expanded feedback and creates a step-by-step CoT plan for what changes are required and how to apply them. 
	\item[iv.] \textbf{Generate Edits}: operator \#4 generates a full revised output in the relevant form (\eg decomposed examples) based on the plan.
\end{itemize}

\subsection{Interactive Collaboration}
\label{subsec:interactive-collaboration}

\begin{figure*}[t!]
	\vspace*{-0.7em}
	\begin{subfigure}[t]{0.9\textwidth}
		\centering
		\captionsetup{justification=centering}
		\includegraphics[scale=0.21]{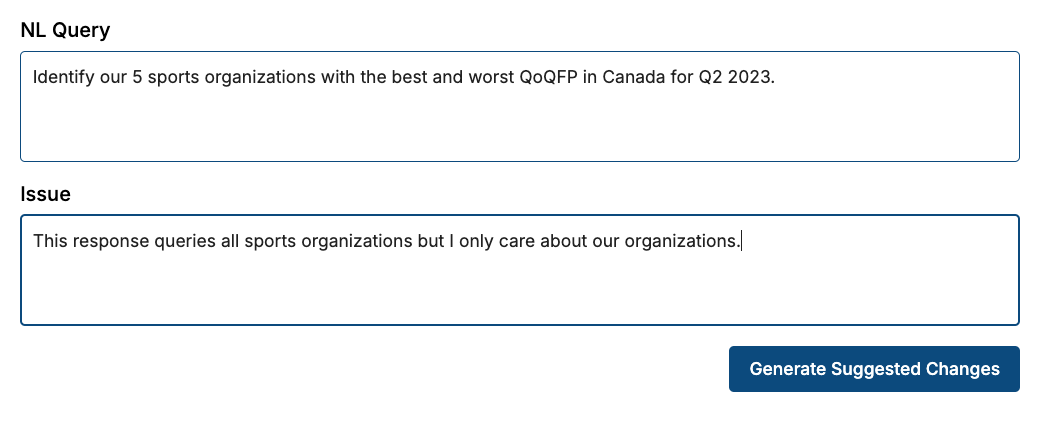}
		\caption{Interface for Submitting Feedback.}
		\label{fig:submitting-feedback}
	\end{subfigure}\vspace*{0.5em}
	\begin{subfigure}[t]{0.22\textwidth}
		\centering
		\captionsetup{justification=centering}
		\includegraphics[scale=0.15]{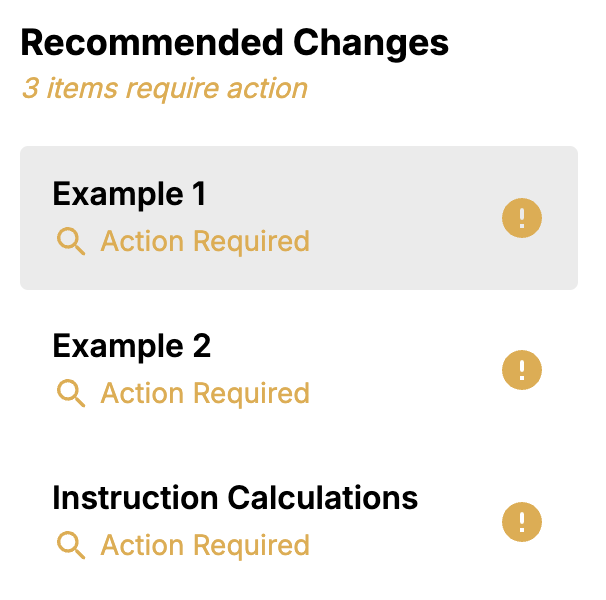}
		\caption{Recommended Edits.}
		\label{fig:recommended-edits}
	\end{subfigure}
	\begin{subfigure}[t]{0.5\textwidth}
		\centering
		\captionsetup{justification=centering}
		\includegraphics[scale=0.12]{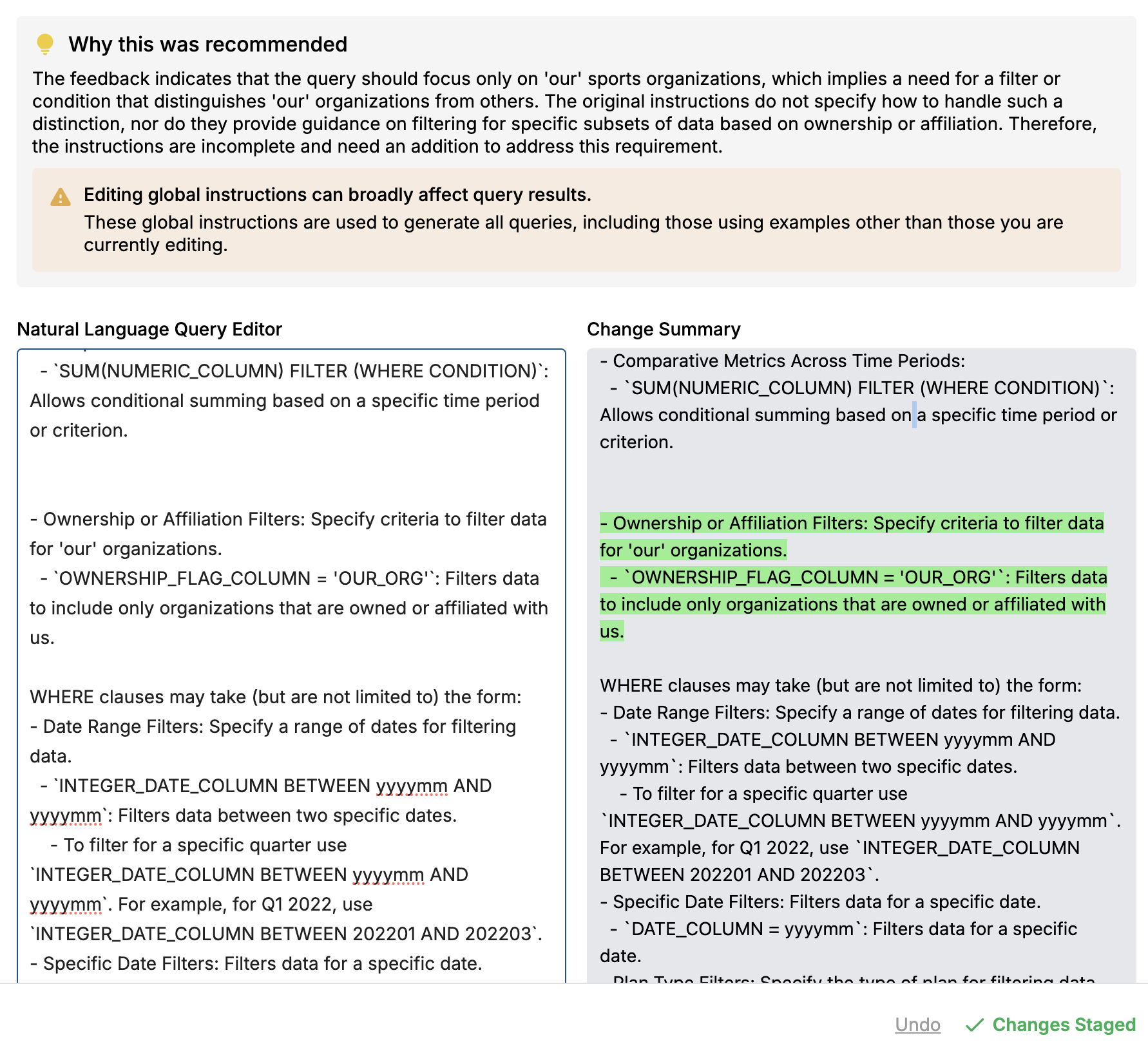}
		\caption{Interface for Reviewing Recommended Edits.}
		\label{fig:reviewing-edits}
	\end{subfigure}
	\begin{subfigure}[t]{0.22\textwidth}
		\centering
		\includegraphics[scale=0.15]{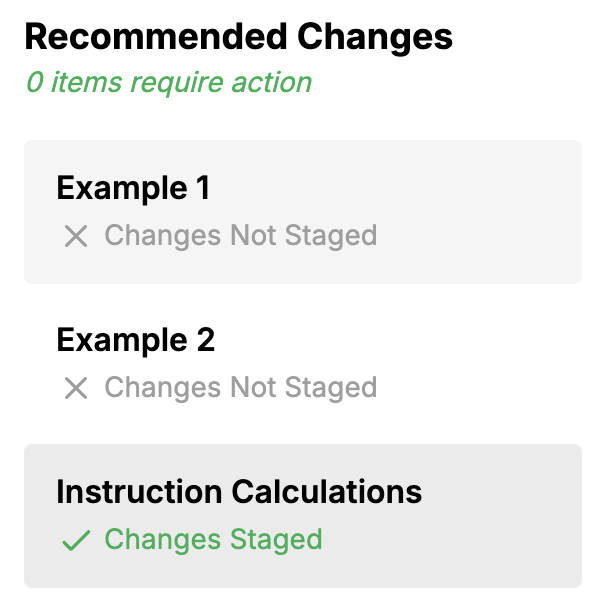}
		\caption{Staged Edits.}
		\label{fig:staged-edits}
	\end{subfigure}\vspace*{0.5em}
	\begin{subfigure}[b]{0.99\textwidth}
		\centering
		\captionsetup{justification=centering}
		\includegraphics[scale=0.12]{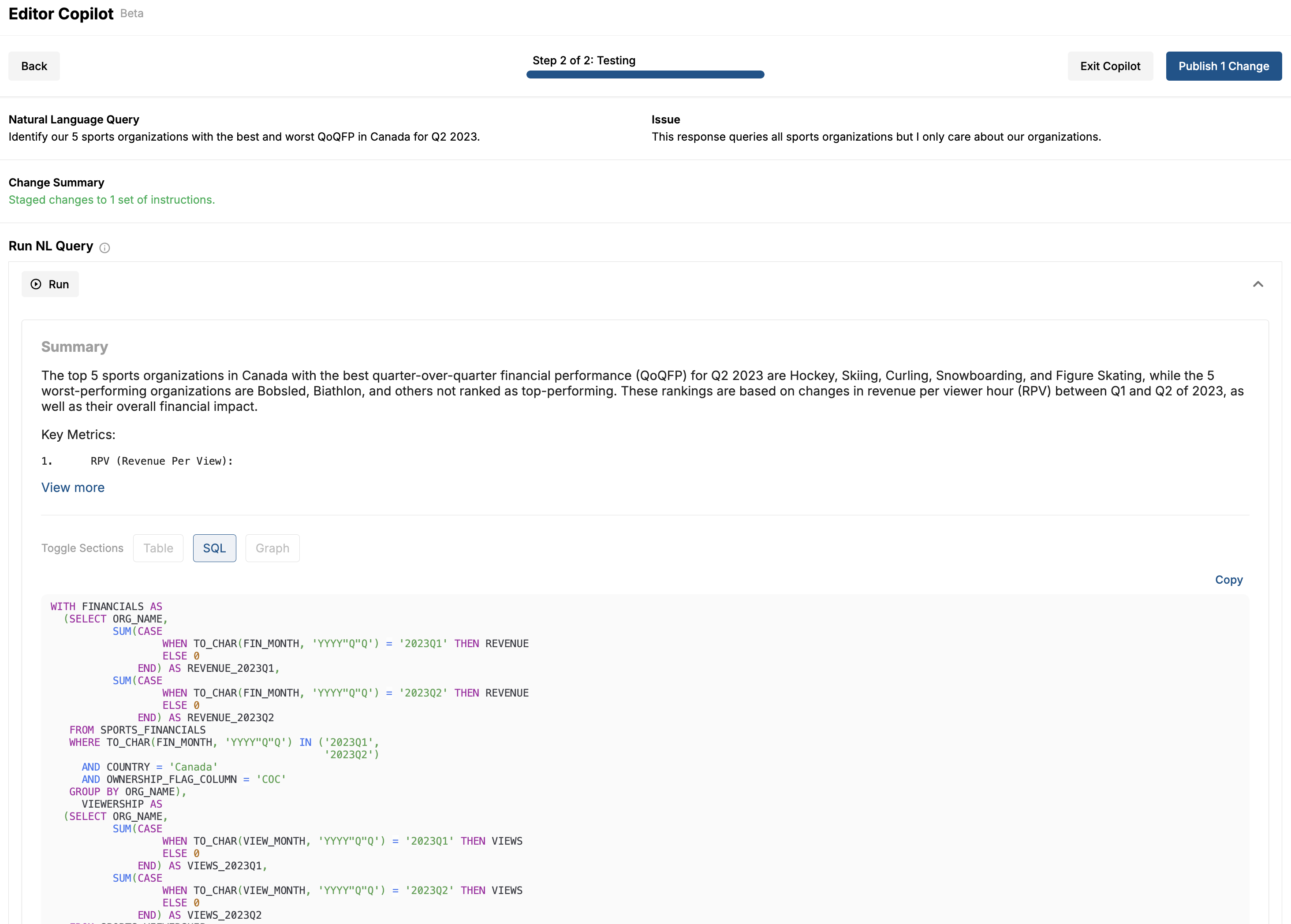}
		\caption{Regeneration Interface including the original natural language query, the feedback, the regenerated query, and\\ summary of the staged edits in the regeneration.}
		\label{fig:regenerating}
	\end{subfigure}
	\centering
	\captionsetup{justification=centering}
	\caption{Example of \genedit~UI interfaces for the .}
	\Description[short description]{long description}
	\label{fig:genedit-ui-screen-examples}
\end{figure*}

\genedit~includes a web-based UI designed to create, manage, and process feedback automatically, leading to edits to examples and instructions in the knowledge set.  
Two key components of the interface are the feedback solver and the knowledge set library. 

\subsubsection{Feedback Solver} 
Once the user submits a natural language query, the interface generates the equivalent SQL with a natural language summary. 
The user can then run the query and inspect the tabular output as well as a set of automatically generated plots. 
If there are any issues with the query, the user can provide natural language feedback as shown in~\fig{fig:submitting-feedback} for $Q_{fin-perf}$ where the feedback is that \emph{`This response queries all sports organizations but I only care about our organizations.'}. 
Once the feedback is submitted, recommended edits (obtained as described above in \sect{subsec:editing-knowledge-set}) are shown in a side panel. 
In our example, the system recommends three edits, two for examples and one for an instruction as shown in~\fig{fig:reviewing-edits}.   
The user can then review each of the edits, make further changes if necessary, and finally stage some or all of them. 
Staging in this case means accepting the edit and taking it to an environment that mimics the deployed system for testing. 
The user staged edits are highlighted in the UI as shown in~\fig{fig:regenerating}. 
Once ready, the user can regenerate the query and continue iterating on their feedback and staging edits. 
The full view containing the various elements after regeneration is shown in~\fig{fig:regenerating}. 
Once done, the edits are submitted and then go through regression testing. If they pass, they are pending for approval. 


\begin{figure*}[t!]
	\centering
	\captionsetup{justification=centering}
	\includegraphics[scale=0.14]{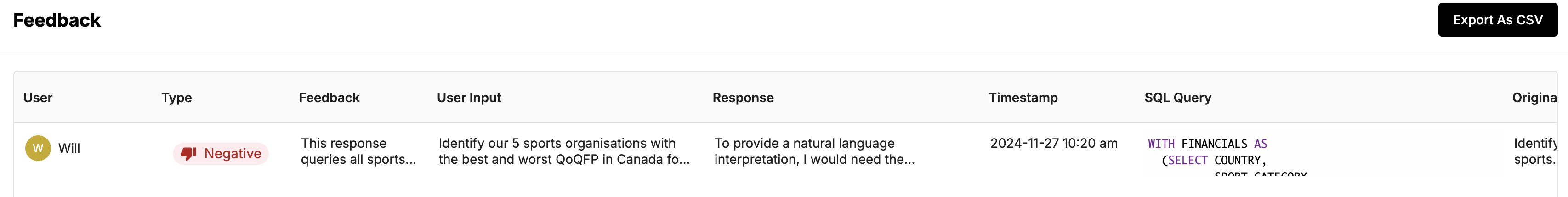}
	\caption{Example of a feedback displayed within the knowledge set library of \genedit.}
	\Description[short description]{long description}
	\label{fig:genedit-lib}
\end{figure*}

\subsubsection{Knowledge Set Library} 
This interface allows an expert user to interact with the knowledge set. 
The library showcases the components of the knowledge set and their provenance, offering full visibility for reversion, comparison, and systematic learning from prior feedback. \fig{fig:genedit-lib} shows an example of how old feedback is shown in a list ordered by timestamp.
Experts may also directly edit the knowledge set within the library outside of the context of a query and its feedback. 

\subsubsection{Evaluation} We evaluate \genedit's edits recommendation module in production based on two metrics: 
i) \emph{how many suggested edits are automatically accepted as is}; 
ii) \emph{how many edits are accepted after re-using the solver or doing manual knowledge set edits?} 

\section{Related Work} 
\label{sec:related-work}

\textbf{Text-to-SQL.} Early approaches struggled to incorporate external knowledge in a flexible manner due to contextual limitations~\cite{zhong2017seq2sql}. 
Relevant schema elements were identified with semantic parsers \cite{yu2018syntaxsqlnet}, queries were generated through sketches \cite{xu2017sqlnet}, and self-correction was conducted through execution-guided decoding \cite{wang2018robust}. 
Conversely, modern LLM-based approaches benefit from their ability to seamlessly integrate relevant knowledge at the query generation stage, leading to improved accuracy and reliability across various stages, \eg schema linking, query generation, and self-correction \cite{gao2023texttosql, lee2024mcssql}.\\


\noindent\textbf{Crowdsourcing Systems.} \genedit~is a crowdsourcing system (CS) from a problem-solving standpoint as it enlists a crowd of humans to help solve a problem defined by the system owners (in this case improving Text-to-SQL generation)~\cite{DBLP:journals/cacm/DoanRH11}. 
Within \genedit, users implicitly collaborate as we piggyback on an existing analytics engine. 
We attract feedback through \emph{instant gratification}, by immediately showing a user their regenerated query to show that their contribution makes a difference. 
CSes were very popular around a decade ago ($\sim$2010s) with many systems proposed~\cite{10.1145/1989323.1989331,DBLP:conf/cidr/ParameswaranP11}. These are similar in that the systems have operations requiring crowds. They also differ in that our approach is meant for improving system primitives instead of supporting an actual computation. 
Recently, a proposal for \emph{prompt engineering} via declarative crowdsourcing has been proposed~\cite{DBLP:conf/cidr/ParameswaranSAJ24}. 
The proposal focuses on providing feedback on LLM-based data transformations to improve cost. \genedit~differs in that it is a generation pipeline; however, our work can be extended in similar ways by getting feedback on latency or specifying a dollar cost and parametrizing \genedit~pipelines differently.\\

\noindent\textbf{Interactive Machine Learning Systems.} 
\genedit~follows the setup of interactive ML systems and has similar goals in removing the ML experts from the feedback loop~\cite{DBLP:journals/aim/AmershiCKK14}. While our UI design is driven by feedback from users of the system, there has been a lot of prior work exploring the best approach to ML system interactivity. Such work highlights the impact of interactions on user behaviour and vice-versa and aims to design interactions based on understood behaviour~\cite{workshop-on-interactive-ml-2013}.
We are only aware of one interactive system for Text-to-SQL that allows users to directly edit a step-by-step explanation of the query generation to fix errors~\cite{DBLP:conf/emnlp/TianZNLK023}. Unlike \genedit, this approach fixes a single instance of a SQL query after generation without continuous improvement, \ie there is no deep integration of the feedback. As such, the same query can fail again later. \genedit edits the knowledge set and hence future prompts to ensure lasting improvements.

\section{Demo}
\label{sec:demo}

We will demonstrate the effectiveness of \genedit~in generating, reviewing, and updating examples and instructions by running queries against a benchmark dataset of choice. 
First, we will take natural language queries and generate equivalent SQL. 
Second, we will identify issues with the generated output and provide feedback through the Feedback Solver interface (shown in \fig{fig:genedit-ui-screen-examples}). 
We will then update the feedback till the query is regenerated to satisfaction. 
Third, we submit the feedback, showcase an accuracy to check for regressions on few selected golden queries and move to human review where we can check the edits through the knowledge set interface (shown in \fig{fig:genedit-lib}). 
This is the interface that users can use to learn from past feedback or move between various knowledge set checkpoints and revert changes.  
Finally, we close the loop by accepting these changes and validating that the previously incorrect queries now return correct results. 


\section{Conclusion}
\label{sec:conclusion}

We introduced \genedit, a collaborative Text-to-SQL generation system designed to address enterprise-specific challenges. 
\genedit\\can generate very complex SQL queries by leveraging a company-specific knowledge set and a pipeline of compounding operators. 
It improves retrieval of query specific knowledge and minimizes the need for LLM reasoning by decomposing text-to-SQL generation using a novel chain-of-thought plan. Finally, \genedit's feedback interface enables continuous improvement. It allows users to iterate on their feedback regenerating the query until satisfaction. The submitted feedback updates the knowledge set which in turn is tested for regression and then merged. This turn improves future generations as it improves SQL generation prompts. 


\section{Acknowledgements}
\label{sec:ack}

We would like to thank Harshini Jayaram, Will Morley, and the broader team at Distyl AI productionizing \genedit~for numerous useful discussions and help with technical issues.  

\balance
\bibliographystyle{abbrv}
\bibliography{references}

\clearpage

\appendix

\definecolor{dkgreen}{rgb}{0,0.6,0}
\definecolor{gray}{rgb}{0.5,0.5,0.5}
\definecolor{mauve}{rgb}{0.58,0,0.82}
\lstset{language=SQL,
	basicstyle={\small\ttfamily},
	belowskip=3mm,
	breakatwhitespace=true,
	breaklines=true,
	classoffset=0,
	columns=flexible,
	commentstyle=\color{dkgreen},
	framexleftmargin=0.25em,
	frameshape={}{yy}{}{},
	keywordstyle=\color{blue},
	numbers=none,
	numberstyle=\tiny\color{gray},
	showstringspaces=false,
	stringstyle=\color{mauve},
	tabsize=3,
	xleftmargin =1em
}

\onecolumn

\section{Output - SQL Generated by Language Model}
\label{sec:output-sql-generated-by-language-model}

\begin{lstlisting}[language=SQL]
WITH 
FINANCIALS AS (
   SELECT ORG_NAME,
           SUM(CASE WHEN TO_CHAR(FIN_MONTH, 'YYYY"Q"Q') = '2023Q1' THEN REVENUE ELSE 0 END) AS REVENUE_2023Q1,
           SUM(CASE WHEN TO_CHAR(FIN_MONTH, 'YYYY"Q"Q') = '2023Q2' THEN REVENUE ELSE 0 END) AS REVENUE_2023Q2
     FROM SPORTS_FINANCIALS
    WHERE TO_CHAR(FIN_MONTH, 'YYYY"Q"Q') IN ('2023Q1', '2023Q2')
      AND COUNTRY = 'Canada'
      AND OWNERSHIP_FLAG_COLUMN = 'COC'
    GROUP BY ORG_NAME
), 
VIEWERSHIP AS (
   SELECT ORG_NAME,
           SUM(CASE WHEN TO_CHAR(VIEW_MONTH, 'YYYY"Q"Q') = '2023Q1' THEN VIEWS ELSE 0 END) AS VIEWS_2023Q1,
           SUM(CASE WHEN TO_CHAR(VIEW_MONTH, 'YYYY"Q"Q') = '2023Q2' THEN VIEWS ELSE 0 END) AS VIEWS_2023Q2
     FROM SPORTS_VIEWERSHIP
    WHERE TO_CHAR(VIEW_MONTH, 'YYYY"Q"Q') IN ('2023Q1', '2023Q2')
      AND COUNTRY = 'Canada'
      AND OWNERSHIP_FLAG_COLUMN = 'COC'
    GROUP BY ORG_NAME
), 
CHANGE_IN_REVENUE AS (
  SELECT
    f.ORG_NAME,
    CAST(f.REVENUE_2023Q2 AS FLOAT) / NULLIF(v.VIEWS_2023Q2, 0) AS RPV,
    CAST(f.REVENUE_2023Q1 AS FLOAT) / NULLIF(v.VIEWS_2023Q1, 0) AS PRIOR_QTR_RPV,
    -1 * (
      (CAST(f.REVENUE_2023Q2 AS FLOAT) / NULLIF(v.VIEWS_2023Q2, 0)) -
      (CAST(f.REVENUE_2023Q1 AS FLOAT) / NULLIF(v.VIEWS_2023Q1, 0))
    ) AS RPV_CHANGE,
    ((CAST(f.REVENUE_2023Q2 AS FLOAT) / NULLIF(v.VIEWS_2023Q2, 0)) - 
     (CAST(f.REVENUE_2023Q1 AS FLOAT) / NULLIF(v.VIEWS_2023Q1, 0)))
    ) * NULLIF(v.VIEWS_2023Q2, 0) as IMPACT,
    ROW_NUMBER() OVER (PARTITION BY f.COUNTRY ORDER BY (-1 * (
    	(CAST(f.REVENUE_2023Q2 AS FLOAT) / NULLIF(v.VIEWS_2023Q2, 0)) - 
    	(CAST(f.REVENUE_2023Q1 AS FLOAT) / NULLIF(v.VIEWS_2023Q1, 0)))
    ) DESC) AS SPORT_RANK,
    ROW_NUMBER() OVER (PARTITION BY f.COUNTRY ORDER BY (-1 * (
    	(CAST(f.REVENUE_2023Q2 AS FLOAT) / NULLIF(v.VIEWS_2023Q2, 0)) -
    	(CAST(f.REVENUE_2023Q1 AS FLOAT) / NULLIF(v.VIEWS_2023Q1, 0)))
    ) ASC) AS WORST_SPORT_RANK
  FROM FINANCIALS f 
   JOIN VIEWERSHIP v ON f.ORG_NAME = v.ORG_NAME
)

SELECT
  SPORT_RANK, ORG_NAME, RPV, PRIOR_QTR_RPV, RPV_CHANGE, IMPACT
FROM
  CHANGE_IN_REVENUE
WHERE
  SPORT_RANK <= 5 OR WORST_SPORT_RANK <= 5
ORDER BY
  SPORT_RANK;
\end{lstlisting}

\end{document}